# Pentagon-Match (PMatch): Identification of View-Invariant Planar Feature for Local Feature Matching-Based Homography Estimation

Yueh-Cheng Huang, Chen-Tao Hsu, and Jen-Hui Chuang
Dept. of Computer Science, National Yang Ming Chiao Tung Univ., Hsinchu, Taiwan

## Abstract

*In computer vision, finding correct point correspondence among images plays an important role in many applications, such as image stitching, image retrieval, visual localization, etc. Most of the research works focus on the matching of local feature before a sampling method is employed, such as RANSAC, to verify initial matching results via repeated fitting of certain global transformation among the images. However, incorrect matches may still exist. Thus, a novel sampling scheme, Pentagon-Match (PMatch), is proposed in this work to verify the correctness of initially matched keypoints using pentagons randomly sampled from them. By ensuring shape and location of these pentagons are view-invariant with various evaluations of cross-ratio (CR), incorrect matches of keypoint can be identified easily with homography estimated from correctly matched pentagons. Experimental results show that highly accurate estimation of homography can be obtained efficiently for planar scenes of the HPatches dataset, based on keypoint matching results provided by LoFTR. Besides, accurate outlier identification for the above matching results and possible extension of the approach for multi-plane situation are also demonstrated.*

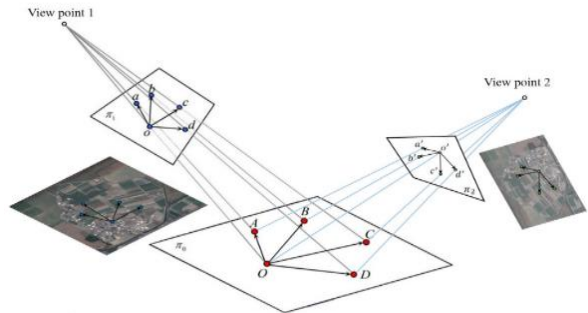

Fig. 1. **Geometry for computing the view-invariant cross-ratio** with respect to a vertex (O) of a pentagon, which can be used, together with CRs associated with the rest four vertices, to identify correct correspondences of pentagonal region in stereo images (see Equations (1) and (2)).

## 1. Introduction

In computer vision, identification of feature points correspondence between two images plays an important role in many applications, such as image stitching, image retrieval, visual localization, etc. Major steps of local feature matching, similar to that described in [1], include: (1) identify image keypoints using selected local feature descriptors, (2) establish initial point correspondences by matching their local features, (3) perform inlier screening (outlier removal) to obtain correct correspondences based on certain geometric transformation between two images. Accordingly, classical algorithms for feature descriptors, such as SIFT [2], SURF [3], ORB [4], GLOH [5], and BOLD [6], use local image characteristics to extract most unique and stable keypoints for images from different views, whereas different algorithms are developed, with various heuristics [7-10], to match them across multiple images via correspondence verification.

On the other hand, most learning-based works of local feature matching are proposed in less than five years for effective detection and description of local image feature [12-16]. Other approaches learn to identify outliers in the matches via various consistency checks, e.g., for pose estimation, by learned inlier classifiers [17-20]. More recently, some state-of-the-art matching results are obtained with SuperGlue [1], a learnable middle-end matcher, between feature extraction and pose estimation, which has an end-to-end architecture consisting of (i) a graph network with alternating self- and cross-attention layers, based on appearance and keypoint location, and (ii) a matching layer. Inspired by [1], a detector-free feature matching method, LoFTR [21], is developed using an efficient variant of the attention layers in Transformer for dense matches in low-texture areas, with better scene estimations, e.g., for homography, obtained in the final stage of correspondence verification. In this paper, ideas similar to self- and cross-attention are employed in the development of a novel way of correspondence verification, based on keypoint location only, for further improvements in homography estimation.

In general, as indicated in [22], there are two main types of algorithms for correspondence verification, i.e., geometric verification methods and the learning-based methods, with only few of the latter are solely based on leaning, such as outlier classification [17, 23] and deep fundamental matrix estimation from data [18]. On the other hand, geometric verification methods identify correct correspondences, and remove outliers, by fitting between-image transformation, e.g., similarity, affinity, homography, or with essential and fundamental matrices, to all initial point matches using sampling methods based on RANSAC (RANdom SAmple Consensus) [24] and its variants. While some of the variants consider different ways of sampling

[25-27], many of them are developed to improve the quality of the fitted model [28-35] and/or the efficiency of the most time-consuming process of quality evaluation [36-38]. While all these methods may involve the matching of planar features, the homography relating a planar surface in different images is not always considered explicitly.

In this paper, a novel planar (pentagonal) model is proposed for the correspondence verification of initially matched local image features for homography estimation. Unlike the geometric verification methods, or other non-RANSAC spatial verification schemes [39-41], one crucial attribute of the pentagonal model is its view-invariant shape description [42] based on cross-ratios (CRs) associated with its vertices, as shown in Fig. 1 for one of the five vertices (randomly selected from initially matched keypoints). Therefore, the model quality is not an issue, as a model can be discarded directly if it is *view-variant*. Furthermore, an outlier in keypoint matching can be identified easily with the underlying homographic transformation associated with the planar region defined by the corresponding pentagon(s).

Specifically, given initial point correspondences from local feature matching between two images ($I_1$ and $I_2$), each partitioned into $N \times N$ blocks, the main steps of the proposed approach to correspondence verification, as well as the estimation(s) of the associated homography, include:

(i) **Pentagon match:** For a pentagonal region arbitrarily selected in each block of $I_1$, check its counterpart in $I_2$ for a correct shape match using five CRs.
(ii) **Homography estimation:** Identify planar regions in $I_1$ (and $I_2$), and associated homographies, by aggregating coplanar pentagonal regions correctly matched in (i).
(iii) **Inlier/outlier Identification:** Identify all correctly (and incorrectly) matched keypoints in $I_1$ (and $I_2$) with homographic constraints established in (ii).

Desirable features of such an approach of homography estimation and correspondence verification include:

I. **Accurate and robust**: Stable and accurate verification can be achieved with view-independent thresholds of CRs for precision shape match in (i), and pixel-based tolerances of image location error in (iii).

II. **Simple and efficient**: (i) is more an order of magnitude simpler in computation complexity compared with an RANSAC-based implementation using homography, and treats all vertices equally in the computation.

III. **General**: In principle, the planar region aggregation in (ii) not only works for scenes with a single planar region, and may also work for multi-plane situations.

IV. **Easy to verify**: Instead of using connecting lines, as in Fig. 2 (a), pentagons found in (i) can serve as landmarks to facilitate location determination, as in Fig. 2 (b), for correct pentagon/keypoint (red/yellow) correspondence.

## 2. A brief review of RANSAC

For the third step of local feature matching mentioned in Sec. 1, sampling methods based on RANSAC are often employed to perform inlier screening (outlier removal) for initially matched keypoints, wherein the correctness of point correspondences can be verified by fitting certain geometric transformation between two images, such as similarity, affinity, homography, etc. A typical RANSAC-based sampling procedure can be summarized as follows.

*Input:*
  $C_n(x_n, y_n)$: correspondence pairs, $1 \leq n \leq N$.
  $\phi$: a selected between-image transformation.
  $E_T$: threshold of fitting error.
  $K$: maximum number of iterations.

*Output:*
  $W_O$: best parameters of $\phi$.

*Procedure RANSAC:*
Step 1: Find a subset $C_m$ from $C_n$ by random selection.
Step 2: Find a set of parameters $w_m$ of model $\phi$ to fit $C_m$.
Step 3: Fed $C_n$ into $\phi$, and find the fitting error, $E_m$.
Step 4: if $E_m < E_T$, let $W_O = w_m$, and report a successful match.
Step 5: if iteration $K$ is not yet reached, go to Step 1; else, report an unsuccessful match and end the procedure.

In this paper, the widely used homography is selected as the underlying geometric transformation between images of planar scenes. Thus, for the above RANSAC-based procedure, each subset $C_m$ will have *four* pairs of matched keypoints from random selection. To gain the desirable features mentioned in the previous section, the proposed approach will use one more pair of matched keypoints (to form an initially matched pentagon) as discussed next.

## 3. Method

For an overview of Steps (i)-(iii) of the proposed approach listed in Sec. 1, which will be elaborated in Subsections 3.1, 3.2, and 3.3, respectively, Fig. 2 (a) shows an initial set of point correspondences obtained using SIFT and the BF-match [43] from OpenCV 4.5.5.6.[1] First, nine pentagons are randomly selected for a 3×3 image partition before seven of them are verified to have the correct shape match according to (i), as shown in Fig. 2 (b). Next, six (red) *coplanar* pentagons[2] are aggregated to represent a planar region according to (ii). Finally, all correctly matched keypoints (yellow circles) are identified by (iii) based on the homography estimated in (ii).

---

[1] More results obtained for initially matched keypoints based on LoFTR will be provided in Sec. 4.

[2] As the scene has a single planar region, the pair of blue pentagons can easily be identified as an erroneous match from the significant difference in their locations relative to the rest six pentagons in the two images.

Note that the correctness of inlier locations can be verified much easier in Fig. 2 (b) than in Fig. 2 (a) via visual inspection as the correctly matched (red) pentagons can be used as reference landmarks. (The incorrect (blue) match can be removed easily with Step A in Sec. 3.2.) Although the shape matching method in (i) can be replaced with a different method using homography, higher computation costs are needed, as will be discussed in Sec. 3.4.

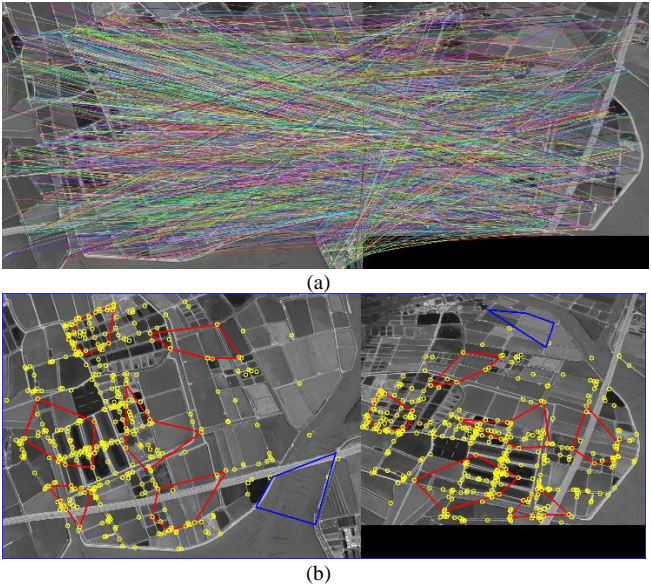

(a)

(b)

Fig. 2. (a) Initial SIFT point correspondences, (b) our easy-to-check point (and region) matching results based on a 3×3 partition of the left image. Red and blue colors correspond to shape matched pentagons, with correct and incorrect locations, respectively.

### 3.1. Pentagon match

To work with two corresponding scenes, say in $I_1$ and $I_2$, which may contain more than one planar regions, one of the stereo images, say $I_1$, is first partitioned into $N \times N$ blocks wherein the shape matching process can be performed for each block for different camera views[3]. To that end, the view-invariant CR is employed in the matching for each selected pentagonal region. For example, with point $O$ being the origin of position vectors of the other four vertices in Fig. 1, two CRs

$$CR_o(a,b,c,d) = \frac{|\vec{a} \times \vec{c}||\vec{b} \times \vec{d}|}{|\vec{b} \times \vec{c}||\vec{a} \times \vec{d}|} \quad (1)$$

and

$$CR_{o\prime}(a',b',c',d') = \frac{|\vec{a'} \times \vec{c'}||\vec{b'} \times \vec{d'}|}{|\vec{b'} \times \vec{c'}||\vec{a'} \times \vec{d'}|} \quad (2)$$

can be defined in View 1 and View 2, respectively. In theory, being invariant under perspective projection, (1) and (2) should be identical. Accordingly, the idea is applied in the following procedure, after an initial local feature matching, for a pair of pentagons established in two different views:

*Procedure Pentagon Match (PMatch):*
1. Randomly select five keypoints in $I_1$, together with their counterparts in $I_2$, from the set of point correspondences obtained from the local feature matching.
2. Compute five pairs of *CR-CR'*, each similar to (1) and (2), for all corresponding vertices in Step 1.
3. Report a successful shape matching if the two values of each *CR-CR'* pair have a negligible difference.
4. Report an unsuccessful match if a predetermined trial count $K_p$ is reached; otherwise, initiate the next trial of pentagon match.

The idea of the above pentagon matching procedure can be regarded as an effective variant of self- (intra-image) and cross- (inter-image) attention [1]. As only the spatial relationship (angular locations) are considered in the pentagon shape model established for a fixed set of five keypoints in each image, the inter-image matching can be accomplished with arbitrary accuracy, resulting superior performance of subsequent estimation of homography and inlier/outlier identification.

For example, cross-ratios $CR_o$ and $CR'_{o\prime}$ in Fig. 1 will be regarded as having a good match by Step 3 if

$$\frac{|CR_o - CR_{o\prime}|}{|CR_o + CR_{o\prime}|} \leq CR_{TH} \quad (3)$$

where the view-independent threshold $CR_{TH}$ can be selected according to desired accuracy in relative angular locations among selected keypoints, as well as the overall pentagon shape, which is set to 5% in our implementation.

Fig. 2 (b) shows some shape matched (red and blue) pentagonal regions thus obtained for a trial count of $K_p = 1,000$[4]. The number of planar regions in the scene will then be determined by merging the above matched pentagons into groups, each corresponds to one planar region.

### 3.2. Homography estimation(s)

Major steps of merging the matched pentagons obtained in the previous subsection into coplanar regions in $I_1$ and $I_2$ can be summarized in the following steps:

A. Identify pairs of mistakenly matched pentagonal regions and remove them from subsequent processes.
B. Incrementally merge the remaining pentagonal regions in $I_1$ (and $I_2$) into a planar group, also using *CR*s as before.

---

[3] In theory, for a scene having only a single planar region, the whole image can be used directly without any partition, and the matching will be performed in (i) for just a single image block. However, an $N \times N$ partition may result in better homography estimation in (ii), as will be demonstrated in the experiments, since a larger convex hull is more likely to be obtained from multiple matched (red) pentagons than a single pentagon.

[4] While more trials may result in more matches, only a single pentagon pair is needed in theory for identifying correct point correspondences in each planar region, as discussed in Sec. 3.3.

C. Establish additional planar groups for pentagons that cannot be merged into existing planar groups.
D. End when all matched pentagon pairs are processed.

Consider shape matched pentagon pairs for the planar scene shown in Fig. 2 (b), one (blue) pair corresponds to erroneous matches, e.g., with unmatched size and location, and will be removed from the final (red) pentagon pairs by Step A. One can see from this typical case that as $I_1$ is partitioned with a 3×3 grid, erroneous matches can be detected easily based on rather simple rules of geometry (instead of considering their precise locations as in Sec. 3.3). For example, for these two images of a planar scene, the 3×3 adjacency relationship among the pentagons in $I_2$ of Fig. 2 should be identical to that among their counterparts in $I_1$[5].

For the example shown in Fig. 2, a single planar region is then established with Step B by merging the remaining six good matches, with the final merged result depicted with a single (red) color. As a matter of fact, Step B corresponds to a shape-matching procedure similar to that presented in Sec. 2.1. In particular, it is implemented by the following procedure for two pairs of pentagon, say $P_1$ and $P_2$:

✧ Randomly select $m$ and $n$ (for example, $m=3$ and $n=2$) corresponding vertices from $P_1$ ($P_2$), with $m+n=5$, to form a new pentagon pair.
✧ Form a second pair with the remaining vertices of $P_1$ and $P_2$ (optional).
✧ Aggregate $P_1$ and $P_2$ into a coplanar group if each of the above pentagon pairs has matched shape.

While adjacent pentagons can be assembled with the above procedure to extend a planar region incrementally, merging them with an arbitrary order will work as well (as the same planar constraint will be established). With correctly matched keypoints identified for the same planar region, the homography between the two images can be estimated easily, e.g., with the procedure reviewed in Sec. 2 but with $C_m$ having at least *five* pairs of matched keypoints[6].

### 3.3. Inlier and outlier Identification

With the planar regions identified with Steps A~D in Sec. 3.2, the rest correct (and incorrect) correspondences of feature points in $I_1$ and $I_2$ can be identified with respect to the established set of planar (homographic) constraints, one for each planar region, as follows:

1. Establish a homographic transformation between $I_1$ and $I_2$ for each pair of corresponding planar regions.
2. For each pair of initially matched feature points, if their positions are predicted accurately by the homographic transformation obtained above, e.g., within 10 pixels, associate them to the corresponding planar region (as inliers); otherwise, mark them as an incorrect match (outliers).

For example, Fig. 2 (b) shows the correct matches (yellow circles) thus obtained for initially matched keypoints in Fig. 2 (a), which can be verified visually by examining their relative locations using nearby pentagon(s) as reference landmarks. The incorrect, and to be discarded, matches (blue circles in Fig. 3), on the other hand, will need to be examined more carefully for their problems. It is not hard to see that, due to the large number of incorrect matches[7], examining their problems is infeasible using connecting lines like those shown in Fig. 2 (a). Therefore, pointing arrows of different colors are depicted in Fig. 3 to help with the examination of inconsistent planar locations of four pairs of incorrect matches.

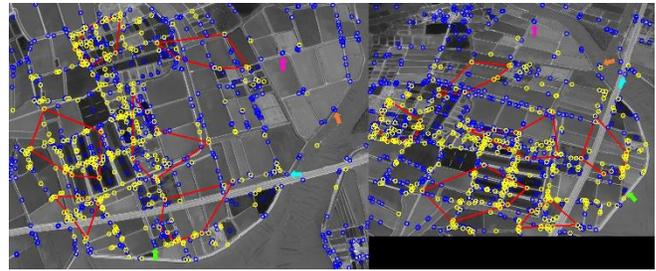

Fig. 3. Identification of correct (same as in Fig. 2) and incorrect (blue) point matches for all initially matched keypoints, with some typical cases of the latter indicated with arrows of different color.

### 3.4. Selection between cross-ratio and homography

In Sec. 3.1 and Sec. 3.2, cross-ratios associated with a pentagon are selected in the computation for (i) efficiency and (ii) robustness. For (i), to determine the planar constraint from four points for the fifth, as in Equation (1), ten multiplications (including four outer products) and one division are needed, while for a similar shape matching scheme using homography, a $(2n \times 9, n = 4)$ matrix will need to be established with 648 (2×4×9×9) multiplications and 72 additions, before nine homography parameters can be found as its eigenvector with $O(n^3)$ in complexity [45], or down to $O(n^2 \sim n^{2.3})$ under certain conditions [46], for $n \geq 4$. As for (ii), a view-invariant threshold, as in Equation (3), can be used, while setting similar threshold for constraint based on homography is not straightforward.

On the other hand, homography is used in Sec. 3.3 for the identification of correct and incorrect matches. This is because only one homography needs to be established for each planar region (in Sec. 3.1 and Sec.3.2, homography needs to be found for each pentagon in each image, and for

---

[5] Other clues for identifying erroneous match include the lack of other keypoints (inliers) associated with such match, as considered in the next subsection, as well as the different color or brightness distributions within the two regions.

[6] Examples for which Step C needs to be performed to established multiple planar regions will be provided in the experiments.

[7] Overall, 557 incorrect matches are identified for the initial 1150 matches found by BF-match [43] for SIFT keypoints obtained by OpenCV 4.5.5.6, showing a serious problem often overlooked by previous research works, as well as the effectiveness of the proposed approach in fixing it.

Table 1. Accuracies of homography estimation on HPatches [46]

| Category | Local feature matching | Homography estimation AUC | | | | | |
| --- | --- | --- | --- | --- | --- | --- | --- |
| | | RANSAC | | | PMatch | | |
| | | @3px | @5px | @10px | @3px | @5px | @10px |
| Detector-based | SIFT+NN | 23.58% | 23.93% | 27.36% | **75.94**% | **79.25**% | **80.80**% |
| | (CPU time)* | (91.04s) | (92.44s) | (93.92s) | (**24.85**s) | (**25.83**s) | (**24.05**s) |
| Detector-free | LoFTR** | 70.81% | 78.61% | 86.19% | **85.46**% | **92.15**% | **95.14**% |
| | (CPU time)* | (432.81s) | (422.07s) | (433.35s) | (**35.60**s) | (**33.51**s) | (**36.61**s) |

*The results are based on a desktop computer with Intel i7 CPU and 16GB RAM, with average-of-ten values used for RANSAC.
**A confidence score of 0.1 is selected for LoFTR to obtain accuracy values closest to those presented in [21].

each trial), from an arbitrary number of correctly matched keypoints. Consequently, to check an initially matched pair of feature points, its location in $I_2$ (and $I_1$ if needed) will be compared with the one predicted by the homogrphy from $I_1$ to $I_2$ (9 multiplications) according to Step 2 of Sec. 3.3. Although similar check in location can be performed by computing CR twice [8], four reference points from all correctly matched keypoints, and two position vectors from these references points, need to be selected somehow.

## 4. Experiment

In this section, we will first compare results of homography estimation, based on RANSAC and PMatch, for initially keypoint correspondences obtained with SIFT+NN and LoFTR, with parameters of LoFTR network provided by the GitHub demo program [32], for images of the HPatches dataset. The database has 57 image sequences, and each sequence has six images of the same scene, wherein the first image is used to produce the rest images based on randomly generated homography matrices[9]. Thus, for each sequence, the accuracy of homography estimation will be evaluated for a total of five image pairs. In addition, in-depth discussion on possible causes of estimation errors will also be provided, which will mainly consider better point correspondences established by LoFTR.

Next, PMatch results are utilized in inlier/outlier identification for some LoFTR-based point correspondences mentioned above. While high percentages of inliers are identified for most HPatches sequences, some less perfect results are also observed and may need further investigation. Finally, we will demonstrate the possibilities of using PMatch for real-world scenes with multiple planar regions.

### 4.1. Homography estimation

In the first experiment, homography estimation results based on the method proposed in Sec. 3.2 are evaluated on the widely adopted HPatches dataset [46], and compared with related results presented in [21], which is based on local feature matching using the detector-free LoFTR. Table 1 shows some results of homography estimation, in terms of area under the cumulative error curve (AUC), for error thresholds of 3, 5, and 10 pixels for the estimation of keypoint location, and for both RANSAC and PMatch.

In Table 1, only LoFTR, the best local feature matching method according to [21], is selected among learning-based, while the traditional SIFT-based image feature matching is also included to show the superior capability of PMatch, compared to RANSAC, in coping with inferior matching results. Without comparing with other RANSAC variants (as most of them focus on best model selection), ten sets of baseline-RANSAC results, with $E_T$ = 10 pixels and $K$ = 1000, are generated for each case, before the best set is selected and presented in Table 1.

As shown in the upper part of Table 1, dramatic improvements up to 55.32% (from 23.93% to 79.25%) over the best-of-ten RANSAC results can be achieved for SIFT-based feature matching by the proposed PMatch procedure, with $K_p$ = 1000, for all three error thresholds. The main reason of such improvements is the prerequisite to the success of RANSAC-based sampling methods that, in statistics, good samples should have larger population than the bad ones, while PMatch only needs to find a minimum of one good (shape verified) pentagon, which can then be used as a reference for accurate localization of planar keypoints.

Regarding the local feature matching of the detector-free LoFTR, as shown in the lower part of Table 1, a confidence score of 0.1 is first selected so that the homography estimation results obtained with our implementation will have values most similar to those presented in [21], i.e., 65.9%, 75.6%, and 84.6%, for LoFTR+RANSAC. It is readily observable that improvements of up to 13% (from 78.61% to 92.15%) can be achieved with LoFTR+PMatch.

On the other hand, one can see that PMatch-based estimation is also much more efficient than that based on RANSAC, as suggested in Sec. 3.4, with the latter highly dependent on the amount of LoFTR matches. (At one time, SIFT+NN and LoFTR generate 171,461 and 908,772 matches, respectively. Please see Supplementary Materials

---
[8] For example, one can see that the location of $o$ in view 1 of Fig. 1 can be determined by $-\vec{a}$ and $-\vec{b}$, each obtained by computing a cross-ratio.

[9] These matrices are also available from the HPatches dataset, and provide the ground truth homography for correctness evaluation.

for more detailed figures for individual HPatches sequence.) As the accuracy of SIFT-based estimations is much worse than that based on LoFTR, possible causes of error in the estimation of homography are mainly discussed below for LoFTR, which include, from major to minor ones:

✧ Visual characteristics of image sequences of HPathces dataset, which may affect the feasibility of estimation.
✧ Parameter settings, e.g., for confidence level in LoFTR, which may affect the quality of image feature matching.
✧ Inherent image quantization errors in location of each shape-matched pentagon found by PMatch.

**Visual characteristics for feasible estimation**

While accuracies of homography estimation are provided in Table 1 for all 57 sequences of HPatches as a whole, examining such accuracy for each of the 57 sequences, as shown in Fig. 4 for LoFTR matching results, will be helpful for better understanding of causes of erroneous estimations, as well as possibilities of future improvements. In general, as one may expect, the PMatch results are mostly better than the best-of-ten RANSAC estimates, as the randomness nature of RANSAC is effectively suppressed by PMatch via view-invariant shape matching of reference pentagons.

Nonetheless, the estimation accuracies shown in Fig. 4 are highly dependent on specific image sequences under consideration for both RANSAC and PMatch. In particular, Fig. 5 shows poor matching results obtained with LoFTR for three image sequences (image pairs from Sequences 7, 17 and 21), wherein good matches are simply not enough for a fair estimation of homography because of undesirable image characteristics.

Table 2. Homography estimation for different confidence scores in LoFTR

| Confidence score | Homography estimation AUC | | | | | |
|---|---|---|---|---|---|---|
| | RANSAC | | | PMatch | | |
| | @3px | @5px | @10px | @3px | @5px | @10px |
| 0.3 | 77.25% | 83.76% | 90.13% | **88.56%** | **94.19**% | **96.48**% |
| 0.5 | 80.28% | 85.20% | 93.30% | **89.01**% | **96.00**% | **97.94**% |
| 0.7 | 82.35% | 90.26% | 95.46% | **93.88**% | **96.07**% | **98.64**% |
| 0.9 | 85.86% | 90.97% | 95.49% | **94.27**% | **96.66**% | **98.77**% |

**Setting different confidence levels for LoFTR**

While a value of 0.1 is chosen as the confidence score for LoFTR in Table 1 so that accuracies of LoFTR+RANSAC in homography estimation will be comparable to those presented in [21], larger values of the confidence score can also be used to achieve better estimations, as shown in Table 2. (This is true only if problems depicted in Fig. 5 will not arise for such stricter confidence requirements of LoFTR.)

One the other hand, it is also readily observable from Table 2 that, compared with PMatch, larger improvements in estimation accuracy, e.g., by increasing the confidence score by 0.2 for LoFTR, can be obtained for RANSAC, except for three (red) cases. This is because, with the view-invariant constraint for shape matched pentagonal regions, PMatch already provides a higher level of confidence.

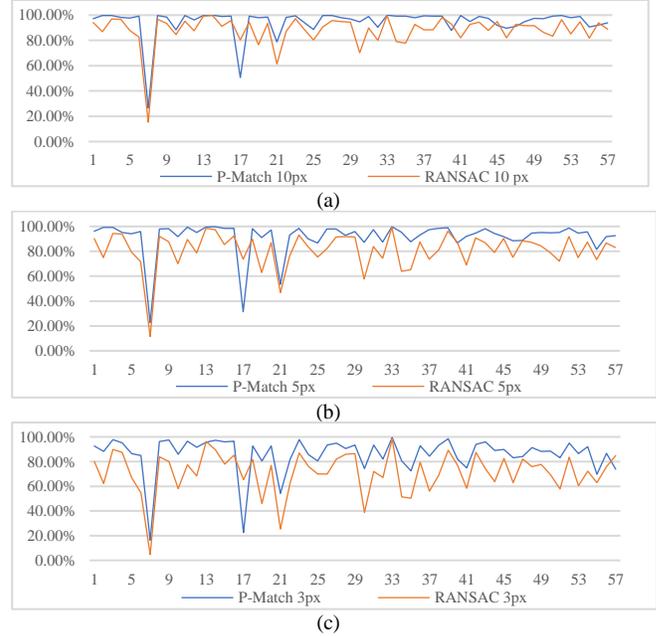

Fig. 4. Accuracies of homography estimation obtained for all 57 sequences of HPatches considered in Table 1, for AUC thresholds of (a) 10 pixels, (b) 5 pixels, and (c) 3 pixels.

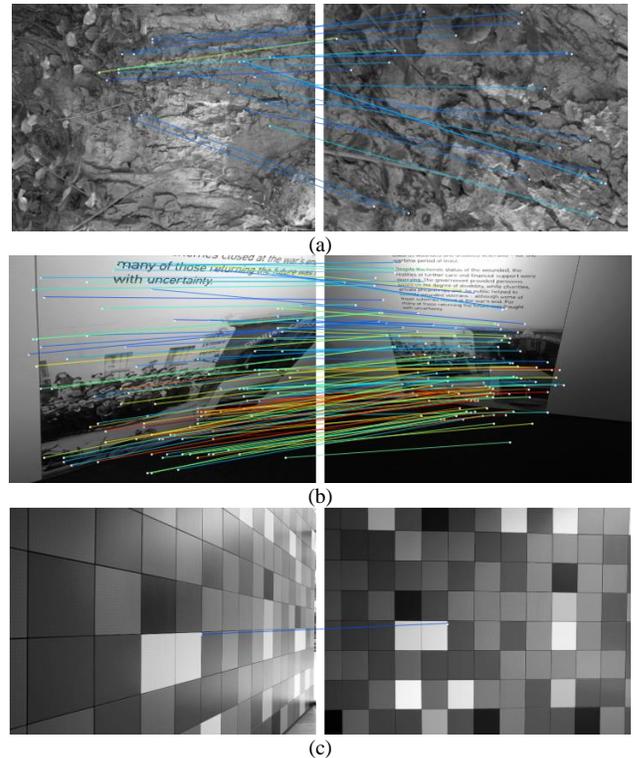

Fig. 5. LoFTR matches for image pairs selected from (a) Sequence 7, (b) Sequence 17, and (c) Sequence 21 of HPatches. While mostly erroneous matches are obtained in (a) and (b) due to low contrast and small area visible from both images, respectively, only a single match is found in (c) because (few) keypoints are having similar local characteristics.

**Quantization of pentagon locations found by PMatch**

Due to the quantization of image coordinates (in pixels), vertices of shape-matched pentagons found by PMatch are located with inherent sub-pixel errors, which will in turn affect the accuracy of subsequent estimation of homography. The problem will be more serious if these selected vertices are confined to a small region. For example, Fig. 6 shows the estimation accuracies depicted in Fig. 5 (c) together with areas of convex hull of such vertices for all 57 sequences of HPatches. One can see that the smallest (< 20% image size) convex hulls of pentagon vertices found by PMatch actually result in the top three drops in estimation accuracy.

In fact, the quantization errors will also affect the accuracy of some *best case* estimation results. For example, Fig. 7 shows three incorrectly matched keypoints for an image pair of Sequence 44, with the (red) keypoint in the right image just having a location error of 11 pixels, which is due to the amplification of quantization error for image features in the near range.

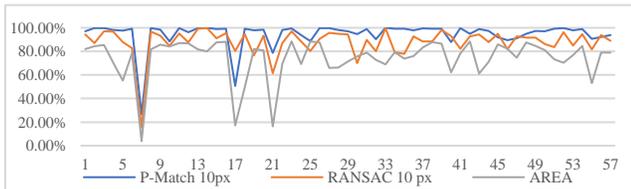

Fig. 6. Three smallest convex hulls of pentagon vertices result in top three drops in estimation accuracy.

### 4.2. Inlier/Outlier Identification

While it is verified in the previous subsection that much higher accuracy of homography estimation can be achieved with proposed PMatch than RANSAC based on the ground truth provided in HPatches, the possibility of using PMatch results for inlier/outlier identification for LoFTR matches is considered in this subsection, which is especially important in practice as the ground truth is usually unavailable.

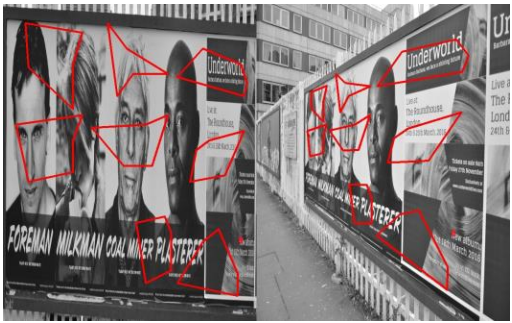

Fig. 7. Two *incorrectly* matched keypoints (see text).

Specifically, let $H_G$ be the ground truth of homography matrix under consideration and $H_P$ be the matrix estimated with PMatch [10]. In addition to showing (red) keypoints which are identified as inliers by $H_P$ but not by $H_G$, as shown in Fig. 8 based on a location accuracy of 10 pixels, three more cases can be depicted: (i) (yellow) keypoints identified as inliers by both $H_P$ and $H_G$, (ii) (blue) keypoints identified as outliers by both $H_P$ and $H_G$, and (iii) (purple) keypoints which are identified as outliers by $H_P$ but not $H_G$.

Figs. 8-10 show examples of some identification results for Sequences 12, 44, and 40, respectively. One can see that as $H_P$ is very accurate, with only a single red point identified in Fig. 10. (Keypoints of case (iii) seldom occur, with only few identified in Fig. 11, and can be found together with extra examples in the Supplementary Materials.) Similar to the red points shown in Fig. 7, such results are mainly due to quantization of image coordinates, and will not cause much problem in the estimation of homography in general.

Besides, due to the superiority in the matching of image features, LoFTR matches often have very high percentages of (yellow) inliers, e.g., 96.2% (1158/1203), 98.1% (2355/2399), and 94.8% (1822/1920) for Figs. 8-10, respectively. Compared to using a large number of connecting lines for pairs of LoFTR-matched keypoints, as shown in Figs. 2 (a) and 5, problems associated with outliers (blue) can be observed more easily using nearby pentagons, which will be more helpful for investigations of further improvement of matching process similar to LoFTR.

On the other hand, Fig. 11 shows an example with only 36.3% (117/325) inliers[11] which can be verified easily using shape matched pentagons, and utilized by the proposed approach to generate accurate estimation of homography. However, the large number of outliers will also hinder the visual examination of these incorrect matches, and other ways of evaluation may need to be explored in the future for possible improvements of the underlying keypoint matching process.

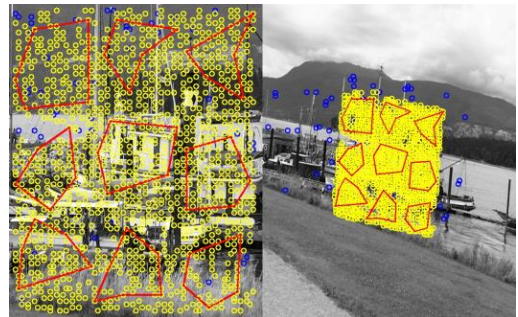

Fig. 8. Inlier/outlier identification for Sequence 12, wherein most outliers correspond to "out of the region" matches.

---

[10] While only $H_G$ is used in Sec. 4.1 to generate ground truth location, e.g., for each keypoint in *left* image of Fig. 7, keypoints matched by LoFTR in *both* images need to be analyzed in this subsection.

[11] Additional statistics can be found in the Supplementary Materials.

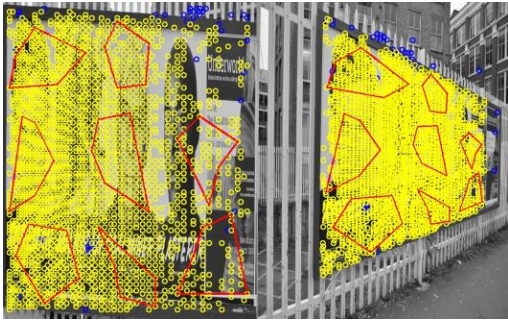

Fig. 9. Inlier/outlier identification for Sequence 44, wherein most outliers correspond to "shifted" matches.

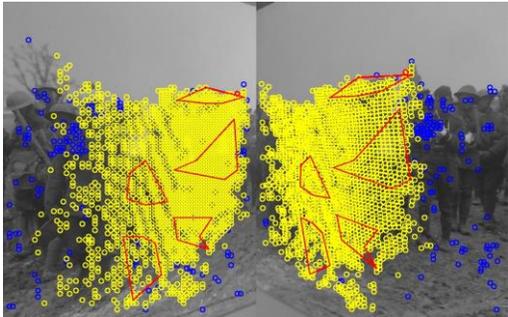

Fig. 10. Inlier/outlier identification for Sequence 40, wherein most outliers correspond to "rotated" matches.

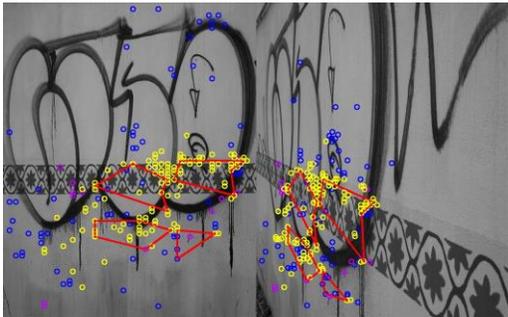

Fig. 11. Inlier/outlier identification for Sequence 29 (see text).

### 4.3. Extension for multi-plane scenes

While the above experimental results demonstrate the effectiveness of the proposed PMatch-based approach to validate correspondences of keypoints in real scenes with a single planar region, as in HPatches dataset, the approach can also be extended to multi-plane situations, based-on Step C of the procedure presented in Sec. 2.2. Fig. 12 shows inlier identification results similar to those obtained for HPatches dataset but for a scene with two planar regions, in two colors (yellow and green) for correctly matched SIFT keypoints. Note that a 2×2 partition (four quadrants) is needed for this example, and further investigation is needed for the selection of appropriate partition[12]. Finally, Fig. 13 shows outlier identification results for incorrectly matched LoFTR keypoints, for a complex scene of a church. The outliers are mainly due to regions away from the near planar façade,[13] which are too small for successful estimations of homography. Further research is needed to address related issues.

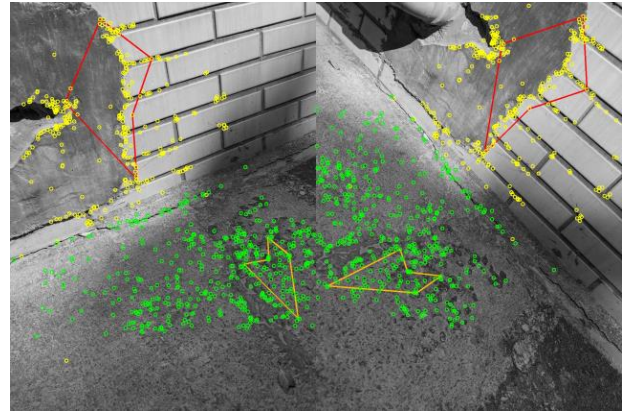

Fig. 12. Inlier identification for a 2-plane scene.

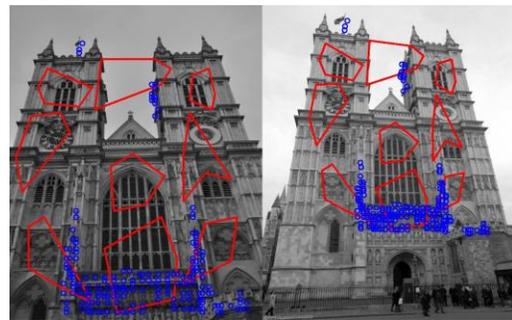

Fig. 13. Inlier/outlier identification identification for a scene with more than two planar regions.

### 5. Conclusions

This study proposes a novel sampling algorithm, Pentagon-Match (PMatch), to verify the correctness of initially matched image keypoints using pentagons randomly sampled from them. By ensuring shape and location of these pentagons are view-invariant with various evaluations of cross-ratio (CR), and deleting them otherwise, incorrect matches of keypoint can be identified easily with homography estimated from correctly matched pentagons. Experimental results show that highly accurate estimation of homography can be obtained efficiently for planar scenes of the HPatches dataset based on keypoint matching results provided by LoFTR[14], with three main factors for better estimation identified for HPatches dataset, LoFTR setting, and PMatch pentagons. In addition, very accurate outlier identification for the above matching results and possible extension of the approach for multi-plane situation are also demonstrated, with both more easily observed using the correctly matched pentagons as references.

---

[12] For HPatches images, PMatch results are actually not sensitive to the way images are partitioned. Results of using some other partitions can be found in the Supplementary Materials

[13] As the vast amount of inliers actually covers a large portion of the church, making visual observation infeasible, related illustrations are provided in the Supplementary Materials.

[14] And with dramatic improvements obtained for SIFT-based matching as well.